%% file: probing.tex
\documentclass[11pt,a4paper]{article}
\usepackage[hyperref]{acl_2018}
\usepackage{times}
\usepackage{latexsym}
\usepackage{graphicx}
\usepackage{amsmath}
\usepackage{array}

\usepackage[utf8]{inputenc}
\usepackage[T1]{fontenc}

\usepackage{url}

\usepackage{paralist}

\aclfinalcopy % Uncomment this line for the final submission
\def\aclpaperid{891} %  Enter the acl Paper ID here

\title{What you can cram into a single \$\&!\#\** vector:\\Probing sentence embeddings for linguistic properties}
\author{
Alexis Conneau \\
Facebook AI Research \\
Universit\'e Le Mans \\
\texttt{aconneau@fb.com} \\
\And
German Kruszewski \\
Facebook AI Research \\
\texttt{germank@fb.com} \\
\And
Guillaume Lample \\
Facebook AI Research \\
Sorbonne Universités \\
\texttt{glample@fb.com} \\
\AND
Loïc Barrault\\
Université Le Mans \\
\texttt{loic.barrault@univ-lemans.fr} \\
\And
Marco Baroni \\
Facebook AI Research \\
\texttt{mbaroni@fb.com} \\
}

\date{}

\begin{document}
\maketitle

\input{tables}

\input{abstract}
\input{introduction}

\input{tasks}

\input{training}
\input{experiments}

\input{related_work}
\input{conclusion}

\section*{Acknowledgments}

We thank David Lopez-Paz, Holger Schwenk, Herv\'{e} J\'{e}gou, Marc'Aurelio Ranzato and Douwe Kiela for useful comments and discussions. 

\bibliography{marco,probing}
\bibliographystyle{acl_natbib}

\input{appendix}

\end{document}

%% file: tables.tex
\newcommand{\insertmlpprobingtable}{
    \begin{table*}[t]
        \resizebox{1\linewidth}{!}{
        \begin{tabular}{@{}l@{\,}|cccccccccc}
        \hline
        \bf Task & \bf SentLen & \bf WC & \bf TreeDepth & \bf TopConst & \bf BShift & \bf Tense & \bf SubjNum & \bf ObjNum &  \bf SOMO & \bf CoordInv  \\
   
        \hline
        \hline
        \multicolumn{11}{l}{\it Baseline representations} \\
        \hline
        Majority vote & 20.0 & 0.5 & 17.9 & 5.0 & 50.0 & 50.0 & 50.0 & 50.0 & 50.0 & 50.0 \\
        Hum. Eval. & 100 & 100 & 84.0 & 84.0 & 98.0 & 85.0 & 88.0 & 86.5 & 81.2 & 85.0 \\
        Length & \bf 100 & 0.2 &	18.1 &	 9.3 &	50.6 &	56.5 &	50.3 &	50.1 &	50.2 &	50.0 \\
        NB-uni-tfidf & 22.7 & \bf 97.8 & 24.1 & 41.9 & 49.5 & 77.7 & 68.9 & 64.0 & 38.0 & 50.5 \\
        NB-bi-tfidf & 23.0 & 95.0 & 24.6 & 53.0 & \bf 63.8 & 75.9 & 69.1 & 65.4 & 39.9 & \bf 55.7 \\
        BoV-fastText & 66.6 & 91.6 & \bf 37.1 & \bf 68.1 & 50.8 & \bf 89.1 & \bf 82.1 & \bf 79.8 & \bf 54.2 & 54.8 \\
        \hline
        \hline
        \multicolumn{11}{l}{\it BiLSTM-last encoder} \\
        \hline
        Untrained & 36.7  & 43.8 &  28.5 & 76.3 & 49.8 & 84.9 & 84.7 & 74.7 & 51.1 & 64.3 \\
        AutoEncoder & \bf 99.3 & 23.3 & 35.6 & 78.2 & 62.0 & 84.3 & 84.7 & 82.1 & 49.9 & 65.1 \\
        NMT En-Fr & 83.5 &  \bf 55.6 &  42.4 & 81.6 & 62.3 & 88.1 & 89.7 & 89.5 & 52.0 & 71.2 \\
        NMT En-De & 83.8 &  53.1 &  42.1 & 81.8 & 60.6 & 88.6 & 89.3 & 87.3 & 51.5 & \bf 71.3 \\
        NMT En-Fi & 82.4 &  52.6 &  40.8 & 81.3 & 58.8 & 88.4 & 86.8 & 85.3 & 52.1 & 71.0 \\
        Seq2Tree & 94.0 & 14.0 & \bf 59.6 & \bf 89.4 & \bf 78.6 & \bf 89.9 & \bf 94.4 & \bf 94.7 & 49.6 & 67.8 \\
        SkipThought & 68.1 & 35.9 &  33.5 & 75.4 & 60.1 & 89.1 & 80.5 & 77.1 & \bf 55.6 & 67.7 \\
        NLI & 75.9 &  47.3 &  32.7 & 70.5 & 54.5 & 79.7 & 79.3 & 71.3 & 53.3 & 66.5 \\
        \hline
        \hline
        \multicolumn{11}{l}{\it BiLSTM-max encoder} \\
        \hline
        Untrained & 73.3 & \bf 88.8 & 46.2 & 71.8 & 70.6 & 89.2 & 85.8 & 81.9 & 73.3 & 68.3 \\
        AutoEncoder & \bf 99.1 & 17.5 & 45.5 & 74.9 & 71.9 & 86.4 & 87.0 & 83.5 & 73.4 & 71.7\\
        NMT En-Fr & 80.1 & 58.3 & 51.7 & 81.9 & 73.7 & 89.5 & 90.3 & 89.1 & 73.2 & 75.4 \\
        NMT En-De & 79.9 & 56.0 & 52.3 & 82.2 & 72.1 & 90.5 & 90.9 & 89.5 & 73.4 & \bf 76.2 \\
        NMT En-Fi & 78.5 & 58.3 & 50.9 & 82.5 & 71.7 & 90.0 & 90.3 & 88.0 & 73.2 & 75.4\\
        Seq2Tree & 93.3 & 10.3 & \bf 63.8 & \bf 89.6 & \bf 82.1 & \bf 90.9 & \bf 95.1 & \bf 95.1 & 73.2 & 71.9 \\
        SkipThought & 66.0 & 35.7 & 44.6 & 72.5 & 73.8 & 90.3 & 85.0 & 80.6 & \bf 73.6 & 71.0 \\
        NLI & 71.7 & 87.3 & 41.6 & 70.5 & 65.1 & 86.7 & 80.7 & 80.3 & 62.1 & 66.8 \\
        \hline
        \hline
        \multicolumn{11}{l}{\it GatedConvNet encoder} \\
        \hline
        Untrained & 90.3 & 17.1  & 30.3  & 47.5 & 62.0 & 78.2 & 72.2 & 70.9 & 61.4 & 59.6 \\
        AutoEncoder & \bf 99.4 & 16.8 & 46.3 & 75.2 & 71.9 & 87.7 & 88.5 & 86.5 & \bf 73.5 & 72.4 \\
        NMT En-Fr & 84.8 &  41.3 &  44.6 & 77.6 & 67.9 & 87.9 & 88.8 & 86.6 & 66.1 & 72.0 \\
        NMT En-De & 89.6 &  49.0 &  50.5 & 81.7 & 72.3 & 90.4 & 91.4 & 89.7 & 72.8 & \bf 75.1 \\
        NMT En-Fi & 89.3 &  \bf 51.5 &  49.6 & 81.8 & 70.9 & 90.4 & 90.9 & 89.4 & 72.4 & \bf 75.1 \\
        Seq2Tree & 96.5 & 8.7 & \bf 62.0 & \bf 88.9 & \bf 83.6 & \bf 91.5 & \bf 94.5 & \bf 94.3 & \bf 73.5 & 73.8 \\
        SkipThought & 79.1 & 48.4 & 45.7 & 79.2 & 73.4 & 90.7 & 86.6 & 81.7 & 72.4 & 72.3 \\
        NLI & 73.8 &  29.2 &  43.2 & 63.9 & 70.7 & 81.3 & 77.5 & 74.4 & 73.3 & 71.0 \\
        
        \hline
        \end{tabular}
        }
        \caption{{\bf Probing task accuracies.} Classification performed by a MLP with sigmoid nonlinearity, taking pre-learned sentence embeddings as input (see Appendix for details and logistic regression results).
        \label{table:probing_results}
        }
        \vspace{-0.4cm}
    \end{table*}
}

\newcommand{\insertlogregprobingtable}{
    \begin{table*}[t]
        \resizebox{1\linewidth}{!}{
        \begin{tabular}{@{}l@{\,}|cccccccccc}
        \hline
        \bf Task & \bf SentLen & \bf WC & \bf TreeDepth & \bf TopConst & \bf BShift & \bf Tense & \bf SubjNum & \bf ObjNum &  \bf SOMO & \bf CoordInv  \\
   
        \hline
        \hline
        \multicolumn{11}{l}{\it Baseline representations} \\
        \hline
        Majority vote & 20.0 & 0.5 & 17.9 & 5.0 & 50.0 & 50.0 & 50.0 & 50.0 & 50.0 & 50.0 \\
        Hum. Eval. & 100 & 100 & 84.0 & 84.0 & 98.0 & 85.0 & 88.0 & 86.5 & 81.2 & 85.0 \\
        Length & 100 & 0.2 &	18.1 &	 9.3 &	50.6 &	56.5 &	50.3 &	50.1 &	50.2 &	50.0 \\
        NB-uni-tfidf & 22.7 & 97.8 & 24.1 & 41.9 & 49.5 & 77.7 & 68.9 & 64.0 & 38.0 & 50.5 \\
        NB-bi-tfidf & 23.0 & 95.0 & 24.6 & 53.0 & 63.8 & 75.9 & 69.1 & 65.4 & 39.9 & 55.7 \\
        BoV fastText & 54.8 & 91.6 & 32.3 & 63.1 & 50.8 & 87.8 & 81.9 & 79.3 & 50.3 & 52.7 \\
        \hline
        \hline
        \multicolumn{11}{l}{\it BiLSTM-last encoder} \\
        \hline
          Untrained & 32.6 & 43.8 & 24.6 & 74.1 & 52.2 & 83.7 & 82.8 & 71.8 & 49.9 & 64.5 \\
          AutoEncoder & 98.9 & 23.3 & 28.2 & 72.5 & 60.1 & 80.0 & 81.2 & 76.8 & 50.7 & 62.5 \\
          NMT En-Fr & 82.9 & 55.6 & 35.8 & 79.8 & 59.6 & 86.0 & 87.6 & 85.5 & 50.3 & 66.1 \\
          NMT En-De & 82.7 & 53.1 & 35.2 & 80.1 & 58.3 & 86.6 & 88.3 & 84.5 & 50.5 & 66.1 \\
          NMT En-Fi & 81.7 & 52.6 & 35.2 & 79.3 & 57.5 & 86.5 & 84.4 & 82.6 & 50.5 & 65.9 \\
          Seq2Tree & 93.2 & 14.0 & 46.4 & 88.5 & 74.9 & 87.3 & 90.5 & 89.7 & 50.6 & 63.4 \\
          SkipThought & 59.5 & 35.9 & 30.2 & 73.1 & 58.4 & 88.7 & 78.4 & 76.4 & 53.0 & 64.6 \\
          NLI & 71.6 & 47.3 & 28.4 & 67.4 & 53.3 & 77.3 & 76.6 & 69.6 & 51.6 & 64.7 \\
        \hline
        \hline
        \multicolumn{11}{l}{\it BiLSTM-max encoder} \\
        \hline
          Untrained & 66.2 & 88.8 & 43.1 & 68.8 & 70.3 & 88.7 & 84.6 & 81.7 & 73.0 & 69.1 \\
          AutoEncoder & 98.5 & 17.5 & 42.3 & 71.0 & 69.5 & 85.7 & 85.0 & 80.9 & 73.0 & 70.9 \\
          NMT En-Fr & 79.3 & 58.3 & 45.7 & 80.5 & 71.2 & 87.8 & 88.1 & 86.3 & 69.9 & 71.8 \\
          NMT En-De & 78.2 & 56.0 & 46.9 & 81.0 & 69.8 & 89.1 & 89.7 & 87.9 & 71.3 & 73.5 \\
          NMT En-Fi & 77.5 & 58.3 & 45.8 & 80.5 & 69.7 & 88.2 & 88.9 & 86.1 & 71.9 & 72.8 \\
          Seq2Tree & 91.8 & 10.3 & 54.6 & 88.7 & 80.0 & 89.5 & 91.8 & 90.7 & 68.6 & 69.8 \\
          SkipThought & 59.6 & 35.7 & 42.7 & 70.5 & 73.4 & 90.1 & 83.3 & 79.0 & 70.3 & 70.1 \\
          NLI & 65.1 & 87.3 & 38.5 & 67.9 & 63.8 & 86.0 & 78.9 & 78.5 & 59.5 & 64.9 \\
        \hline
        \hline
        \multicolumn{11}{l}{\it GatedConvNet encoder} \\
        \hline
          Untrained & 90.3 & 17.1 & 30.3 & 47.5 & 62.0 & 78.2 & 72.2 & 70.9 & 61.4 & 59.1 \\
          AutoEncoder & 99.3 & 16.8 & 41.9 & 69.6 & 68.1 & 85.4 & 85.4 & 82.1 & 69.8 & 70.6 \\
          NMT En-Fr & 84.3 & 41.3 & 36.9 & 73.8 & 63.7 & 85.6 & 85.7 & 83.8 & 58.8 & 68.1 \\
          NMT En-De & 87.6 & 49.0 & 44.7 & 78.8 & 68.8 & 89.5 & 89.6 & 86.8 & 69.5 & 70.0 \\
          NMT En-Fi & 89.1 & 51.5 & 44.1 & 78.6 & 67.2 & 88.7 & 88.5 & 86.3 & 68.3 & 71.0 \\
          Seq2Tree & 94.5 &  8.7 & 53.1 & 87.4 & 80.9 & 89.6 & 91.5 & 90.8 & 68.3 & 71.6 \\
          SkipThought & 73.2 & 48.4 & 40.4 & 76.2 & 71.6 & 89.8 & 84.0 & 79.8 & 68.9 & 68.0 \\
          NLI & 70.9 & 29.2 & 38.8 & 59.3 & 66.8 & 80.1 & 77.7 & 72.8 & 69.0 & 69.1 \\
        
        \hline
        \end{tabular}
        }
        \caption{{\bf Probing task accuracies with Logistic Regression.} Taking pre-learned sentence embeddings as input.
        \label{table:probing_logreg_results}
        }
    \end{table*}
}

\newcommand{\insertdownstreamtable}{
    \begin{table*}[t]
        \resizebox{1\linewidth}{!}{
        \begin{tabular}{@{}l@{\,}|ccccccccccc}
        \hline
        \bf Model & \bf MR & \bf CR & \bf SUBJ & \bf MPQA & \bf SST-2 & \bf SST-5 & \bf TREC & \bf MRPC &  \bf SICK-E & \bf SICK-R & \bf STSB \\

        \hline
        \hline
        \multicolumn{10}{l}{\it Baseline representations} \\
        \hline
        	Chance & 50.0 & 63.8 & 50.0 & 68.8 & 50.0 & 28.6 & 21.2 & 66.5 & 56.7 & 0.0 & 0.0 \\
			BoV fastText & 78.2 & 80.2 & 91.8 & 88.0 & 82.3 & 45.1 & 83.4 & 74.4 & 78.9 & 82.0 & 70.2 \\
        \hline
        \hline
        \multicolumn{10}{l}{\it BiLSTM-last encoder} \\
        \hline
          Untrained & 69.7 & 70.2 & 84.8 & 87.0 & 77.2 & 37.6 & 79.6 & 68.5 & 71.6 & 68.2 & 54.8 \\
          AutoEncoder & 66.0 & 70.7 & 85.7 & 81.1 & 70.0 & 36.2 & 84.0 & 69.9 & 72.2 & 67.6 & 58.3 \\
          NMT En-Fr & 74.5 & 78.7 & 90.3 & 88.9 & 79.5 & 42.0 & 91.2 & 73.7 & 79.7 & 78.3 & 69.9 \\
          NMT En-De & 74.8 & 78.4 & 89.8 & 88.7 & 78.8 & 42.3 & 88.0 & 74.1 & 78.8 & 77.5 & 69.3 \\
          NMT En-Fi & 74.2 & 78.0 & 89.6 & 88.9 & 78.4 & 39.6 & 84.6 & 75.6 & 79.1 & 77.1 & 67.1 \\
          Seq2Tree & 62.5 & 69.3 & 85.7 & 78.7 & 64.4 & 33.0 & 86.4 & 73.6 & 71.9 & 59.1 & 44.8 \\
          SkipThought & 77.1 & 78.9 & 92.2 & 86.7 & 81.3 & 43.9 & 82.4 & 72.7 & 77.8 & 80.0 & 73.9 \\
          NLI & 77.3 & 84.1 & 88.1 & 88.6 & 81.7 & 43.9 & 86.0 & 74.8 & 83.9 & 85.6 & 74.2 \\

        \hline
        \hline
        \multicolumn{10}{l}{\it BiLSTM-max encoder} \\
        \hline
          Untrained & 75.6 & 78.2 & 90.0 & 88.1 & 79.9 & 39.1 & 80.6 & 72.2 & 80.8 & 83.3 & 70.2 \\
          AutoEncoder & 68.3 & 74.0 & 87.2 & 84.6 & 70.8 & 34.0 & 85.0 & 71.0 & 75.3 & 70.4 & 55.1 \\
          NMT En-Fr & 76.5 & 81.1 & 91.4 & 89.7 & 77.7 & 42.2 & 89.6 & 75.1 & 79.3 & 78.8 & 68.8 \\
          NMT En-De & 77.7 & 81.2 & 92.0 & 89.7 & 79.3 & 41.0 & 88.2 & 76.2 & 81.0 & 80.0 & 68.7 \\
          NMT En-Fi & 77.0 & 81.1 & 91.5 & 90.0 & 80.3 & 43.4 & 87.2 & 75.0 & 81.7 & 80.3 & 69.5 \\
          Seq2Tree & 65.2 & 74.4 & 88.3 & 80.2 & 66.5 & 31.6 & 85.0 & 72.0 & 74.8 & 65.1 & 36.1 \\
          SkipThought & 78.0 & 82.8 & 93.0 & 87.3 & 81.5 & 41.9 & 86.8 & 73.2 & 80.0 & 82.0 & 71.5 \\
          NLI & 79.2 & 86.7 & 90.0 & 89.8 & 83.5 & 46.4 & 86.0 & 74.5 & 84.5 & 87.5 & 76.6 \\

        \hline
        \hline
        \multicolumn{10}{l}{\it GatedConvNet encoder} \\
        \hline
          Untrained & 65.5 & 65.3 & 78.3 & 82.9 & 65.8 & 34.0 & 67.6 & 68.1 & 61.6 & 56.7 & 38.9 \\
          AutoEncoder & 72.1 & 74.1 & 86.6 & 86.0 & 74.4 & 36.6 & 79.6 & 69.7 & 72.0 & 65.8 & 45.5 \\
          NMT En-Fr & 74.5 & 78.3 & 88.7 & 88.4 & 76.8 & 38.3 & 86.2 & 72.5 & 77.3 & 73.2 & 60.4 \\
          NMT En-De & 77.1 & 80.4 & 90.9 & 89.2 & 79.2 & 41.9 & 90.4 & 76.8 & 81.9 & 78.7 & 69.4 \\
          NMT En-Fi & 76.9 & 82.0 & 91.2 & 90.0 & 78.8 & 41.9 & 90.0 & 76.7 & 81.1 & 79.5 & 70.8 \\
          Seq2Tree & 65.3 & 73.1 & 85.0 & 79.8 & 63.7 & 31.8 & 81.2 & 72.9 & 74.0 & 58.4 & 30.8 \\
          SkipThought & 76.0 & 81.7 & 91.5 & 87.2 & 77.9 & 41.5 & 88.8 & 72.3 & 79.5 & 80.0 & 67.8 \\
          NLI & 76.7 & 84.7 & 87.4 & 89.1 & 79.2 & 40.9 & 82.0 & 70.8 & 82.0 & 84.7 & 64.4 \\
        \hline
        \hline
% * <alexis.conneau@gmail.com> 2018-02-21T16:37:57.584Z:
%
% ^.
        \multicolumn{10}{l}{\it Other sentence embedding methods} \\
        \hline
          SkipThought & 79.4 & 83.1 & 93.7 & 89.3 & 82.9 & - & 88.4 & 72.4 & 79.5 & 85.8 & 72.1 \\

          InferSent & 81.1 & 86.3 & 92.4 & 90.2 & 84.6 & 46.3 & 88.2 & 76.2 & 86.3 & 88.4 & 75.8 \\

          MultiTask & 82.4 & 88.6 & 93.8 & 90.7 & 85.1 & - & 94.0 & 78.3 & 87.4 & 88.5 & 78.7 \\        
        \hline
        \end{tabular}
        }
        \caption{{\bf Downstream tasks results for various sentence encoder architectures pre-trained in different ways.}
        \label{table:downstream_results}
        }
    \end{table*}
}

\newcommand{\inserttestresults}{
    \begin{table}[h!]
        \resizebox{1\columnwidth}{!}{
        \begin{tabular}{@{}l@{\,}|ccccc}
        \hline
        \bf Model & \bf En-Fr & \bf En-De & \bf En-Fi & \bf Seq2Tree & \bf NLI \\
        \hline
        Gated ConvNet & 25.9 & 17.0 & 14.2 & 52.3 & 83.5 \\
        BiLSTM-last & 27.3 & 17.9 & 14.3 & 55.2 & 84.0 \\ 
        BiLSTM-max & 27.0 & 18.0 & 14.7 & 53.7 & 85.3 \\
        \hline
        BiLSTM-Att & 39.1 & 27.2 & 21.9 & 58.4 & - \\
        \hline
        \end{tabular}
        }
        \caption{{\bf Test results for training tasks.} Figure of merit is BLEU score for NMT and accuracy for Seq2Tree and NLI.
        \label{table:test_results}
        }
    \end{table}
}

\newcommand{\insertembsize}{
\begin{figure}[h!]
  \centering
  \includegraphics[width=1.0\linewidth]{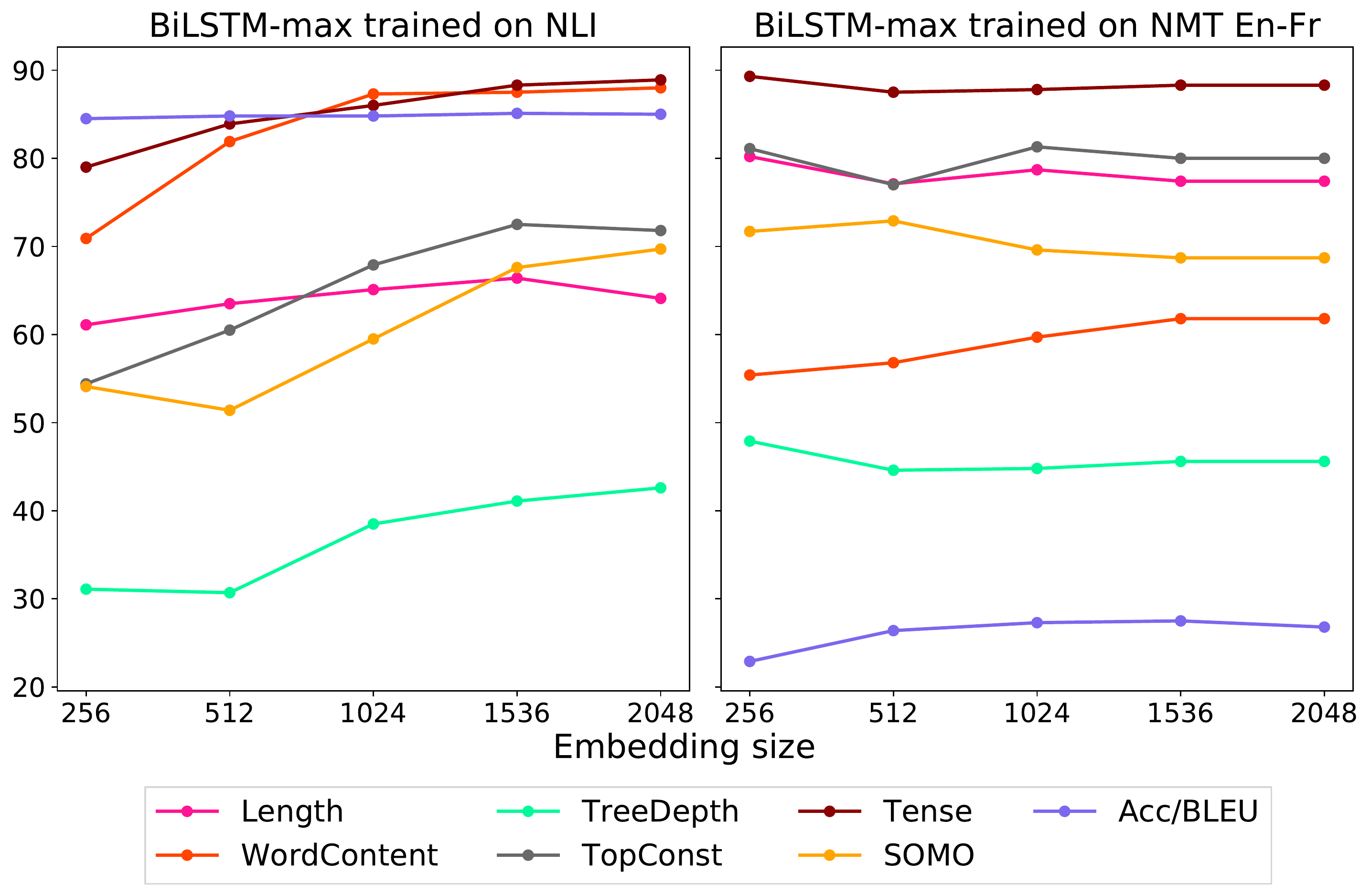}
    \caption{\label{fig:qualitative_trained} {\bf Evolution of probing tasks results wrt.~embedding size}. The sentence representations are generated by a BiLSTM-max encoder trained on either NLI or NMT En-Fr, with increasing sentence embedding size.}
    \label{fig:embsize}
\end{figure}
}

\newcommand{\insertexamples}{
    \begin{table*}[t]
        \resizebox{1\linewidth}{!}{
        \begin{tabular}{|m{2cm}|m{6.2cm}|m{8.4cm}|} 
        \hline
        task & source & target \\
        \hline
        \hline
        AutoEncoder & I myself was out on an island in the Swedish archipelago , at Sandhamn . & I myself was out on an island in the Swedish archipelago , at Sand@ ham@ n . \\
        \hline
        NMT En-Fr & I myself was out on an island in the Swedish archipelago , at Sandhamn . & Je me trouvais ce jour là sur une île de l' archipel suédois , à Sand@ ham@ n .\\
        \hline
        NMT En-De & We really need to up our particular contribution in that regard . & Wir müssen wirklich unsere spezielle Hilfs@ leistung in dieser Hinsicht aufstocken . \\
        \hline
        NMT En-Fi & It is too early to see one system as a universal panacea and dismiss another . & Nyt on liian aikaista nostaa yksi järjestelmä jal@ usta@ lle ja antaa jollekin toiselle huono arvo@ sana . \\
        \hline
        SkipThought & the old sami was gone , and he was a different person now . & the new sami didn 't mind standing barefoot in dirty white , sans ra@ y-@ bans and without beautiful women following his every move .\\
        \hline
        Seq2Tree & Dikoya is a village in Sri Lanka . & $(_{\text{ROOT}}$ $(_S$ $(_{\text{NP}}$ NNP $)_{\text{NP}}$ $(_{\text{VP}}$ VBZ $(_{\text{NP}}$ $(_{\text{NP}}$ DT NN $)_{\text{NP}}$ $(_{\text{PP}}$ IN $(_{\text{NP}}$ NNP NNP $)_{\text{NP}}$ $)_{\text{PP}}$ $)_{\text{NP}}$ $)_{\text{VP}}$ . $)_{\text{S}}$ $)_{\text{ROOT}}$ \\
        \hline
        \end{tabular}
        }
        \caption{\textbf{Source and target examples for seq2seq training tasks.} \label{table:examples}
        }
     \vspace{-0.4cm}
    \end{table*}
}

\newcommand{\insertcorrelationmatrix}{
  \begin{figure}[h!]
    \centering
    \includegraphics[width=1.0\linewidth]{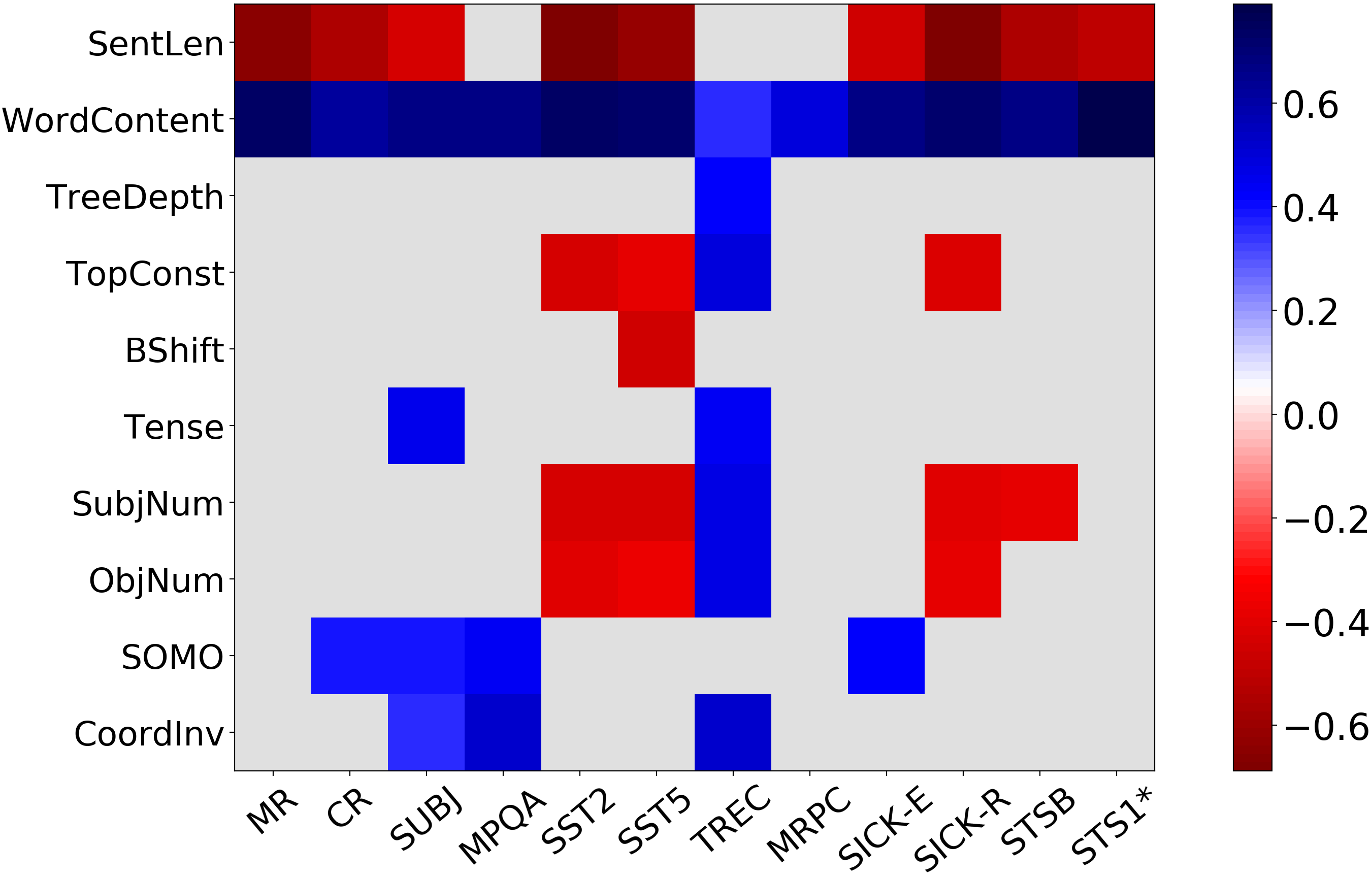}
      \caption{\label{fig:correl_matrix} {\bf Spearman correlation matrix between probing and downstream tasks.} Correlations based on all sentence embeddings we investigated (more than 40). Cells in gray denote task pairs that are not significantly correlated (after correcting for multiple comparisons).}
      \vspace{-0.2cm}
  \end{figure}
}

\newcommand{\insertalongtraining}{
  \begin{figure}[h!]
    \centering
    \includegraphics[width=.99\linewidth]{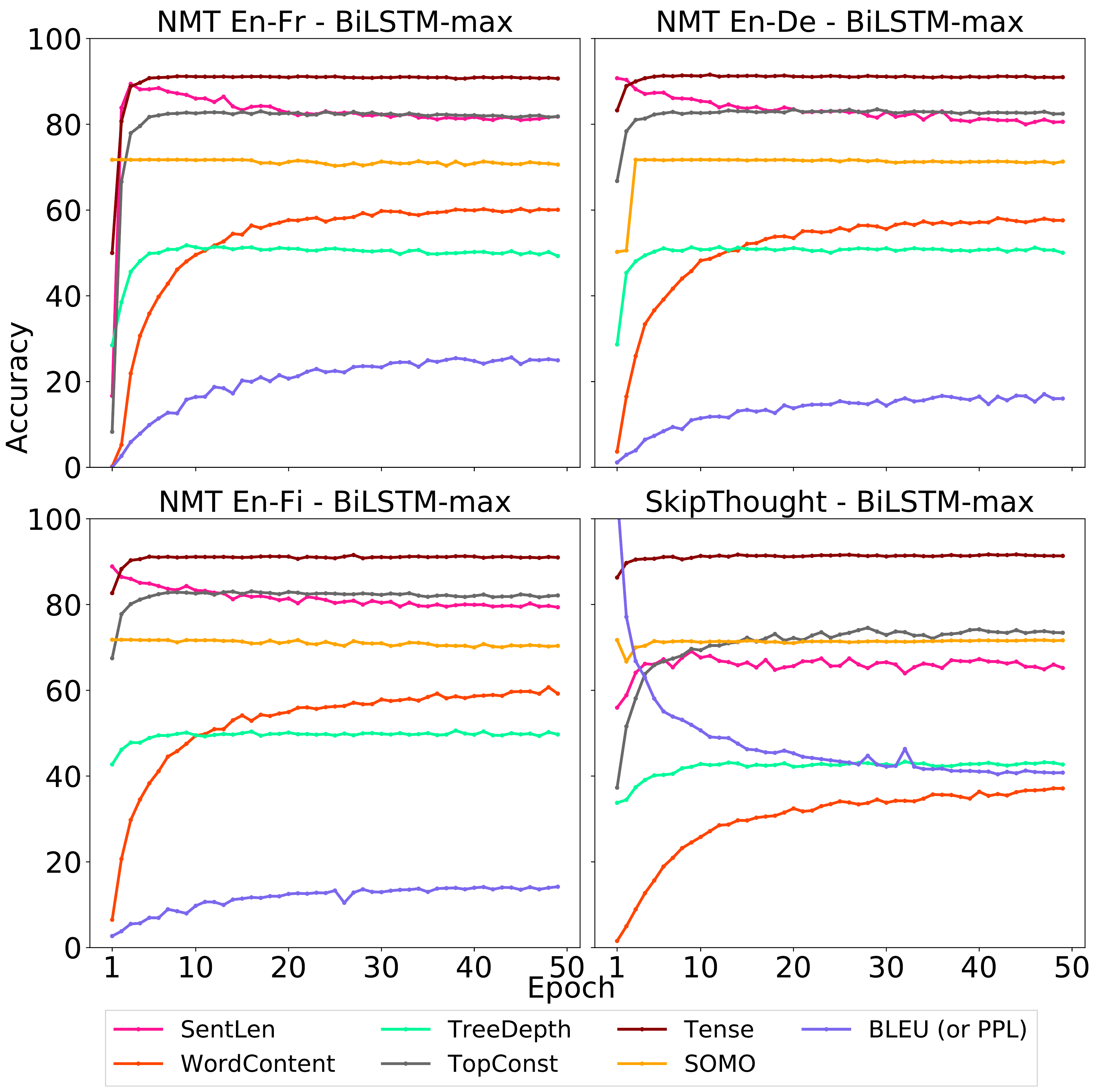}
      \caption{\label{fig:along_training} {\bf Probing task scores after each training epoch, for NMT and SkipThought.} We also report training score evolution: BLEU for NMT; perplexity (PPL) for SkipThought.}
     \vspace{-0.2cm}
  \end{figure}
}

%% file: abstract.tex
\begin{abstract}
Although much effort has recently been devoted to training high-quality sentence embeddings, we still have a poor understanding of what they are capturing. ``Downstream'' tasks, often based on sentence classification, are commonly used to evaluate the quality of sentence representations. The complexity of the tasks makes it however difficult to infer what kind of information is present in the representations. We introduce here 10 probing tasks designed to capture simple linguistic features of sentences, and we use them to study embeddings generated by three different encoders trained in eight distinct ways, uncovering intriguing properties of both encoders and  training methods. 
\end{abstract}

%% file: introduction.tex
\section{Introduction}
\label{sec_introduction}

Despite Ray Mooney's quip that you cannot cram the meaning of a whole
\%\&!\$\# sentence into a single \$\&!\#\** vector, sentence embedding
methods have achieved impressive results in tasks
ranging from machine translation \cite{Sutskever:etal:2014,
  cho2014learning} to entailment detection 
\cite{williams2017broad}, spurring the quest for ``universal
embeddings'' trained once and used in a variety of applications
\cite[e.g.,][]{kiros2015skip,Conneau:etal:2017,subramanian2018learning}. Positive results on
concrete problems suggest that embeddings capture important
linguistic properties of sentences. However, real-life ``downstream''
tasks require complex forms of inference, making it difficult to
pinpoint the information a model is relying
upon. Impressive as it might be that a system can tell that the
sentence ``A movie that doesn't aim too high, but it doesn't need to''
\cite{Pang:Lee:2004} expresses a subjective viewpoint, it is hard
to tell \emph{how} the system (or even a human) comes to this
conclusion. Complex tasks can also carry hidden biases that
models might lock onto \cite{Jabri:etal:2016}. For example,
\newcite{Lai:Hockenmaier:2014} show that the simple heuristic of
checking for explicit negation words leads to good
accuracy in the SICK sentence entailment task.

Model introspection techniques have been applied to sentence encoders
in order to gain a better understanding of which properties of the
input sentences their embeddings retain (see Section
\ref{sec_related}). However, these techniques often depend on the
specifics of an encoder architecture, and consequently cannot be used
to compare different methods. \newcite{Shi:etal:2016} and
\newcite{Adi:etal:2017} introduced a more general approach, relying on
the notion of what we will call \emph{probing tasks}. A probing task
is a classification problem that focuses on simple linguistic
properties of sentences. For example, one such task might require to
categorize sentences by the tense of their main verb. Given an encoder
(e.g., an LSTM) pre-trained on a certain task (e.g., machine
translation), we use the sentence embeddings it produces to train the
tense classifier (without further embedding tuning). If
the classifier succeeds, it means that the pre-trained encoder is
storing readable tense information into the embeddings it creates. Note that: %
\begin{inparaenum}[(i)]
\item The probing task asks a simple question, minimizing interpretability problems.
\item Because of their simplicity, it is easier to control for
  biases in probing tasks than in downstream tasks.
\item The probing task methodology is agnostic with respect to the
  encoder architecture, as long as it produces a vector
  representation of sentences.
\end{inparaenum}

We greatly extend earlier work on probing tasks as follows. First, we introduce a larger set of probing tasks (10 in total),
organized by the type of linguistic properties they
probe. Second, we systematize the probing task methodology,
controlling for a number of possible nuisance factors, and framing all
tasks so that they only require single sentence representations as
input, for maximum generality and to ease result
interpretation. Third, we use our probing tasks to explore a wide
range of state-of-the-art encoding architectures and training methods,
and further relate probing and downstream task performance.  Finally,
we are publicly releasing our probing data sets and tools, hoping they will become a standard way to study the linguistic
properties of sentence embeddings.\footnote{\url{https://github.com/facebookresearch/SentEval/tree/master/data/probing}}

%% file: tasks.tex
\section{Probing tasks}
\label{sec_tasks}

In constructing our probing benchmarks, we adopted the following
criteria. First, for generality and interpretability, the task
classification problem should only require single sentence embeddings
as input (as opposed to, e.g., sentence and word embeddings, or
multiple sentence representations). Second, it should be possible to
construct large training sets in order to train parameter-rich
multi-layer classifiers, in case the relevant properties are
non-linearly encoded in the sentence vectors. Third, nuisance
variables such as lexical cues or sentence length should be controlled
for. Finally, and most importantly, we want tasks that address an
interesting set of linguistic properties. We thus strove to come up
with a set of tasks that, while respecting the previous constraints,
probe a wide range of phenomena, from superficial properties of
sentences such as which words they contain to their hierarchical
structure to subtle facets of semantic acceptability. We think the
current task set is reasonably representative of different linguistic
domains, but we are not claiming that it is exhaustive. We
expect future work to extend it.

The sentences for all our tasks are extracted from the Toronto Book
Corpus \citep{Zhu:etal:2015}, more specifically from the random
pre-processed portion made available by
\newcite{Paperno:etal:2016}. We only sample sentences in the 5-to-28
word range. We parse them with the Stanford Parser (2017-06-09
version), using the pre-trained PCFG model \citep{Klein:Manning:2003},
and we rely on the part-of-speech, constituency and dependency parsing
information provided by this tool where needed. For each task, we
construct training sets containing 100k sentences, and 10k-sentence
validation and test sets. All sets are balanced, having an equal
number of instances of each target class.

\paragraph{Surface information} These tasks test the extent to which
sentence embeddings are preserving surface properties of the sentences
they encode. One can solve the surface tasks by simply looking at
tokens in the input sentences: no linguistic knowledge is called
for. The first task is to predict the \emph{length} of sentences in
terms of number of words (\textbf{SentLen}). Following
\newcite{Adi:etal:2017}, we group sentences into 6 equal-width bins by
length, and treat SentLen as a 6-way classification task. The
\emph{word content} (\textbf{WC}) task tests whether it is possible to
recover information about the original words in the sentence from its
embedding. We picked 1000 mid-frequency words from the source corpus
vocabulary (the words with ranks between 2k and 3k when sorted by
frequency), and sampled equal numbers of sentences that contain one
and only one of these words. The task is to tell which of the 1k words
a sentence contains (1k-way classification). This setup allows us to
probe a sentence embedding for word content without requiring an
auxiliary word embedding (as in the setup of Adi and colleagues).

\paragraph{Syntactic information} The next batch of tasks test whether
sentence embeddings are sensitive to syntactic properties of the
sentences they encode. The \emph{bigram shift} (\textbf{BShift}) task
tests whether an encoder is sensitive to legal word orders. In this
binary classification problem, models must distinguish intact
sentences sampled from the corpus from sentences where we inverted two
random adjacent words (``What \emph{you are} doing out there?'').

The \emph{tree depth} (\textbf{TreeDepth}) task checks whether an
encoder infers the hierarchical structure of
sentences, and in particular whether it can group sentences by the
depth of the longest path from root to any leaf. Since tree depth is naturally correlated with sentence
length, we de-correlate these variables through a structured sampling procedure. In the
resulting data set, tree depth values range from 5 to 12, and the task
is to categorize sentences into the class corresponding to their depth
(8 classes). As an example, the following is a long (22 tokens) but
shallow (max depth: 5) sentence: ``[$_1$ [$_2$ But right now, for the
time being, my past, my fears, and my thoughts [$_3$ were [$_4$ my
[$_5$business]]].]]'' (the outermost brackets correspond to the ROOT
and S nodes in the parse).

In the top constituent task (\textbf{TopConst}), sentences must be
classified in terms of the sequence of top constituents immediately
below the sentence (S) node. An encoder that successfully addresses
this challenge is not only capturing latent syntactic structures, but
clustering them by constituent types. TopConst was introduced by
\newcite{Shi:etal:2016}. Following them, we frame it as a 20-way
classification problem: 19 classes for the most frequent top
constructions, and one for all other constructions. As an example,
``[Then] [very dark gray letters on a black screen] [appeared] [.]''
has top constituent sequence: ``ADVP NP VP .''.

Note that, while we would not expect an untrained human subject to be
explicitly aware of tree depth or top constituency, similar
information must be implicitly computed to correctly parse sentences,
and there is suggestive evidence that the brain tracks something
akin to tree depth during sentence processing \cite{Nelson:etal:2017}.

\paragraph{Semantic information} These tasks also rely on syntactic
structure, but they further require some understanding of what a
sentence denotes. The \textbf{Tense} task asks for the tense of the
main-clause verb (VBP/VBZ forms are labeled as present, VBD as
past). No target form occurs across the train/dev/test split, so that
classifiers cannot rely on specific words (it is not clear that Shi
and colleagues, who introduced this task, controlled for this
factor). The \emph{subject number} (\textbf{SubjNum}) task focuses on
the number of the subject of the main clause (number in English is
more often explicitly marked on nouns than verbs). Again, there is no
target overlap across partitions. Similarly, \emph{object number}
(\textbf{ObjNum}) tests for the number of the direct object of the
main clause (again, avoiding lexical overlap). To solve the previous
tasks correctly, an encoder must not only capture tense and number,
but also extract structural information (about the main clause and its
arguments). We grouped Tense, SubjNum and ObjNum with the semantic
tasks, since, at least for models that treat words as unanalyzed
input units (without access to morphology), they must rely on what a
sentence denotes (e.g., whether the described event took place in the
past), rather than on structural/syntactic information. We recognize,
however, that the boundary between syntactic and semantic tasks is
somewhat arbitrary.

In the \emph{semantic odd man out} (\textbf{SOMO)} task, we modified
sentences by replacing a random noun or verb \emph{o} with another
noun or verb \emph{r}. To make the task more challenging, the bigrams
formed by the replacement with the previous and following words in the
sentence have frequencies that are comparable (on a log-scale) with
those of the original bigrams. That is, if the original sentence
contains bigrams $w_{n-1}o$ and $o w_{n+1}$, the corresponding bigrams
$w_{n-1}r$ and $r w_{n+1}$ in the modified sentence will have
comparable corpus frequencies. No sentence is included in both
original and modified format, and no replacement is repeated across
train/dev/test sets. The task of the classifier is to tell whether a
sentence has been modified or not. An example modified sentence is: ``
No one could see this Hayes and I wanted to know if it was real or a
\emph{spoonful} (orig.: \emph{ploy}).'' Note that judging plausibility
of a syntactically well-formed sentence of this sort will often
require grasping rather subtle semantic factors, ranging
from selectional preference to topical coherence.

The coordination inversion (\textbf{CoordInv}) benchmark contains
sentences made of two coordinate clauses. In half of the sentences, we
inverted the order of the clauses. The task is to tell whether a
sentence is intact or modified. Sentences are balanced in terms of
clause length, and no sentence appears in both original and inverted
versions. As an example, original ``They might be only memories, but I
can still feel each one'' becomes: ``I can still feel each one, but
they might be only memories.'' Often, addressing CoordInv requires an
understanding of broad discourse and pragmatic factors.
\\

Row \textbf{Hum.~Eval.}~of Table \ref{table:probing_results} reports
human-validated ``reasonable'' upper bounds for all the tasks,
estimated in different ways, depending on the tasks. For
the surface ones, there is always a straightforward correct answer
that a human annotator with enough time and patience could find. The
upper bound is thus estimated at 100\%. The TreeDepth, TopConst, Tense, SubjNum
and ObjNum tasks depend on automated PoS and parsing annotation. In these
cases, the upper bound is given by the proportion of
sentences correctly annotated by the automated procedure. To estimate
this quantity, one linguistically-trained author checked the
annotation of 200 randomly sampled test sentences from each
task. Finally, the BShift, SOMO and CoordInv manipulations can
accidentally generate acceptable sentences. For example, one
modified SOMO sentence is: ``He pulled out the large round
\emph{onion} (orig.: \emph{cork}) and saw the amber balm
inside.'', that is arguably not more anomalous than the original. For
these tasks, we ran Amazon Mechanical
Turk experiments in which
subjects were asked to judge whether 1k randomly sampled test
sentences were acceptable or not. Reported human accuracies are based on
majority voting. See Appendix for details.

%% file: training.tex
\section{Sentence embedding models}
\label{sec_training}
In this section, we present the three sentence encoders that we consider and the seven tasks on which we train them.

\insertexamples

\subsection{Sentence encoder architectures}
A wide variety of neural networks encoding sentences into  fixed-size representations exist. We focus here on three that have been shown to perform well on standard NLP tasks.
% % The study of other more advanced encoders is left for future work.

\paragraph{BiLSTM-last/max}
For a sequence of T words $\{ w_{t}\}_{t=1,\ldots,T}$, a bidirectional LSTM computes a set of T vectors $\{h_t\}_t$. For $t\in [1,\ldots, T]$, $h_t$ is the concatenation of a forward LSTM and a backward LSTM that read the sentences in two opposite directions. %
% \begin{eqnarray*}
% \overrightarrow{h_t} &=& \overrightarrow{\text{LSTM}}_t( w_1, \ldots, w_T) \\
% \overleftarrow{h_t} &=& \overleftarrow{\text{LSTM}}_t(w_1, \ldots, w_T) \\
% h_t &=& [\overrightarrow{h_t}, \overleftarrow{h_t}]
% \end{eqnarray*}
%DK: Again, I think we can have clearer notation here.
%AC: done
We experiment with two ways of combining the varying number of
$(h_1, \ldots, h_T)$ to form a fixed-size vector, either by selecting
the last hidden state of $h_T$ or by selecting the maximum value over
each dimension of the hidden units. The choice of these models are
motivated by their demonstrated efficiency  in seq2seq
\cite{Sutskever:etal:2014} and universal sentence representation
learning \cite{Conneau:etal:2017}, respectively.\footnote{We also experimented with
  a unidirectional LSTM, with consistently poorer results.}

\paragraph{Gated ConvNet}
We also consider the non-recurrent convolutional equivalent of LSTMs, based on stacked gated temporal convolutions. Gated convolutional networks were shown to perform well as  neural machine translation encoders \cite{Gehring:etal:2017} and  language modeling decoders \cite{dauphin2016language}. The encoder is composed of an input word embedding table that is augmented with positional encodings \cite{sukhbaatar2015end}, followed by a stack of temporal convolutions with small kernel size. The output of each convolutional layer is filtered by a gating mechanism, similar to the one of LSTMs. Finally,  max-pooling along the temporal dimension is performed on the output feature maps of the last convolution \cite{collobert2008unified}.

\subsection{Training tasks}
% Our aim is to understand what linguistic properties are captured in sentence encoders. To that aim, and to understand the impact of the training task, We learn the three encoder architectures presented above on seven tasks, most of which consist of structure prediction in a sequence to sequence architecture \cite{cho2014learning, sutskever2014sequence}, but also of classification \cite{bowman2015large}.

%\paragraph{Neural Machine Translation}
% TODO: put the following line in the introduction
%\citet{zhou2016deep} showed that, when trained on a large corpus as WMT'14 English-to-French, a sequence to sequence model without attention can obtain a competitive result of 36.3 BLEU score\footnote{while their attention-based model obtains 37.7 BLEU}, which suggests that a fixed-size representation can capture a significant amount of linguistic information.
Seq2seq systems have shown strong results in machine translation \cite{zhou2016deep}. They consist of an \textit{encoder} that encodes a source sentence into a fixed-size representation, and a \textit{decoder} which acts as a conditional language model and that generates the target sentence. We train \textbf{Neural Machine Translation} systems on three language pairs using about 2M sentences from the Europarl corpora \cite{koehn2005europarl}. We pick {\bf English-French}, which involves two similar languages, {\bf English-German}, involving larger syntactic differences, and {\bf English-Finnish}, a distant pair. %
%\paragraph{AutoEncoder}
We also train with an \textbf{AutoEncoder} objective \cite{Socher:etal:2011} on Europarl source English sentences. %
%\paragraph{Sequence to Tree}
Following \newcite{vinyals2015grammar}, we train a seq2seq architecture to generate linearized grammatical parse trees (see Table \ref{table:examples}) from source sentences (\textbf{Seq2Tree}). We use the Stanford parser to generate trees for Europarl source English sentences. %
%\paragraph{SkipThought}
We  train \textbf{SkipThought} vectors \cite{kiros2015skip} by
predicting the next sentence given the current one \cite{tang2017trimming}, on 30M sentences from the Toronto Book Corpus, excluding those in the probing sets.
%\paragraph{Natural Language Inference}
Finally, following \citet{Conneau:etal:2017}, we train sentence encoders on \textbf{Natural Language Inference} using the concatenation of the SNLI \cite{bowman2015large} and MultiNLI \cite{bowman2015large}  data sets (about 1M sentence pairs). In this task, a sentence encoder is trained to encode two sentences, which are fed to a classifier and whose role is to distinguish whether the sentences are contradictory, neutral or entailed. Finally, as in \newcite{Conneau:etal:2017}, we also include \textbf{Untrained} encoders with random weights, which act as random projections of pre-trained word embeddings.

\subsection{Training details}
BiLSTM encoders use 2 layers of 512 hidden units ($\sim$4M parameters), Gated ConvNet has 8 convolutional layers of 512 hidden units, kernel size 3 ($\sim$12M parameters). We use pre-trained fastText word embeddings of size 300 \cite{mikolov:lrec:2018} without fine-tuning, to isolate the impact of encoder architectures and to handle words outside the training sets. Training task performance and further details are in Appendix.

%% file: experiments.tex
\section{Probing task experiments}
\label{sec_experiments}

\insertmlpprobingtable

\paragraph{Baselines} Baseline and human-bound performance are
reported in the top block of Table
\ref{table:probing_results}. \textbf{Length} is a linear classifier
with sentence length as sole feature. \textbf{NB-uni-tfidf} is a
Naive Bayes classifier using words' tfidf scores as features,
\textbf{NB-bi-tfidf} its extension to bigrams. Finally, \textbf{BoV-fastText}
derives sentence representations by averaging the fastText embeddings
of the words they contain (same embeddings used as input to the
encoders).\footnote{Similar results are obtained summing embeddings,
  and using GloVe embeddings \cite{Pennington:etal:2014}.}

Except, trivially, for Length on SentLen and the NB baselines on
WC, there is a healthy gap between top baseline performance
and human upper bounds. NB-uni-tfidf evaluates to what extent our
tasks can be addressed solely based on knowledge about the
distribution of words in the training sentences. Words are of course
to some extent informative for most tasks, leading to relatively high
performance in Tense, SubjNum and ObjNum. Recall that the words
containing the probed features are disjoint between train and test
partitions, so we are not observing a confound here, but
rather the effect of the redundancies one expects in natural language data. For
example, for Tense, since sentences often contain more than one verb in
the same tense, NB-uni-tfidf can exploit
non-target verbs as cues: the NB features most
associated to the past class are verbs in the past tense (e.g ``sensed'', ``lied'', ``announced''), and similarly for present (e.g ``uses'', ``chuckles'', ``frowns''). Using bigram features (NB-bi-tfidf)
brings in general little or no improvement with respect to the unigram
baseline, except, trivially, for the BShift task, where NB-bi-tfidf can easily
detect unlikely bigrams. NB-bi-tfidf has below-random
performance on SOMO, confirming that the semantic intruder is not
given away by superficial bigram cues. %  Importantly, the BoV method is
% already clearly outperforming the Naive-Bayes, with more sophisticated
% encoders further improving the results, which confirms that the
% linguistic knowledge they capture goes beyond superficial word
% distribution statistics.

Our first striking result is the good overall performance of Bag-of-Vectors, confirming early insights that aggregated
word embeddings capture surprising amounts of sentence information
\cite{ThePham:etal:2015,Arora:etal:2017,Adi:etal:2017}.  BoV's good
WC and SentLen performance was already established by
\newcite{Adi:etal:2017}. Not surprisingly, word-order-unaware BoV
performs randomly in BShift and in the more sophisticated semantic
tasks SOMO and CoordInv. More interestingly, BoV is very good at the
Tense, SubjNum, ObjNum, and TopConst tasks (much better than the
word-based baselines), and well above chance in TreeDepth. The good
performance on Tense, SubjNum and ObjNum has a straightforward
explanation we have already hinted at above. Many sentences are
naturally ``redundant'', in the sense that most tensed verbs in a
sentence are in the same tense, and similarly for number in nouns. In
95.2\% Tense, 75.9\% SubjNum and 78.7\% ObjNum test sentences, the
target tense/number feature is also the majority one for the whole
sentence. Word embeddings capture features such as number and tense
\cite{Mikolov:etal:2013a}, so aggregated word embeddings will
naturally track these features' majority values in a sentence. BoV's
TopConst and TreeDepth performance is more surprising. Accuracy is
well above NB, showing that BoV is exploiting cues beyond
specific words strongly associated to the target classes. We conjecture
 that more abstract word features captured by the embeddings (such as the part of speech of a word)
might signal different syntactic structures. For example, sentences in
the ``WHNP SQ .'' top constituent class (e.g., ``\emph{How} long before you leave us again?'') must contain a wh word, and
will often feature an auxiliary or modal verb. BoV can rely on this
information to noisily predict the correct class.

%

% A first striking and general observation from Table \ref{} is that different encoder architectures trained on the same task, with similar performance\footnote{See Appendix for more details on the results on the training task}, can obtain very different results on the probing tasks, coherent with \newcite{Conneau:etal:2017} findings for downstream tasks.

\paragraph{Encoding architectures} Comfortingly, proper encoding
architectures clearly outperform BoV. An interesting observation in Table \ref{table:probing_results} is that different encoder architectures trained with the same objective, and achieving similar performance on the training task,\footnote{See Appendix for details on training task performance.} can lead to linguistically different embeddings, as indicated by the probing tasks. Coherently with the findings of \newcite{Conneau:etal:2017} for the downstream tasks, this suggests that the prior imposed by the encoder architecture strongly preconditions the nature of the embeddings. Complementing recent evidence
that convolutional architectures are on a par with recurrent ones in
seq2seq tasks \cite{Gehring:etal:2017}, we find that Gated ConvNet's
overall probing task performance is comparable to that of the best
LSTM architecture (although, as shown in Appendix, the LSTM has a
slight edge on downstream tasks). We also replicate the finding of
\newcite{Conneau:etal:2017} that BiLSTM-max outperforms BiLSTM-last
both in the downstream tasks (see Appendix) and in the probing
tasks (Table \ref{table:probing_results}). Interestingly, the latter
only outperforms the former in SentLen, a task that captures a
superficial aspect of sentences (how many words they contain), that
could get in the way of inducing more useful linguistic
knowledge.

\paragraph{Training tasks} We focus next on how different training tasks
affect BiLSTM-max, but the patterns are generally
representative across architectures. NMT training leads to encoders that
are more linguistically aware than those trained on the NLI data set,
despite the fact that we confirm the finding of Conneau and colleagues
that NLI is best for downstream tasks (Appendix). Perhaps, NMT
captures richer linguistic features useful for the probing tasks,
whereas shallower or more \emph{ad-hoc} features might help more in
our current downstream tasks. Suggestively, the one task where
NLI clearly outperforms NMT is WC. Thus,
NLI training is better at preserving shallower word features that
might be more useful in downstream tasks
(cf.~Figure \ref{fig:correl_matrix} and discussion
there).%  Interestingly, target language does not seem to have a big
% impact on NMT training, with very similar performance obtained when
% training to translate into very different languages such as French and
% Finnish.

Unsupervised training (SkipThought and AutoEncoder) is not on a par
with supervised tasks, but still effective. AutoEncoder training
leads, unsurprisingly, to a model excelling at SentLen, but it attains
low performance in the WC prediction task. This curious result might indicate that the latter information is stored in the embeddings in a complex way, not easily readable by our MLP. At the other end, Seq2Tree is
trained to predict annotation from the same parser we used to create
some of the probing tasks. Thus, its high performance on TopConst,
Tense, SubjNum, ObjNum and TreeDepth is probably an artifact. Indeed, for most of these tasks, Seq2Tree performance is above the human
bound, that is, Seq2Tree learned to mimic the parser errors
in our benchmarks. For the more challenging SOMO and CoordInv tasks,
that only indirectly rely on tagging/parsing information, Seq2Tree is comparable to NMT, that does not use explicit syntactic
information.

Perhaps most interestingly, BiLSTM-max already achieves very good
performance without any training (Untrained row in Table
\ref{table:probing_results}). Untrained BiLSTM-max also performs quite
well in the downstream tasks (Appendix). This architecture must encode
priors that are intrinsically good for sentence
representations. Untrained BiLSTM-max exploits the input fastText
embeddings, and multiplying the latter by a random recurrent matrix
provides a form of positional encoding. However, good performance in a
task such as SOMO, where BoV fails and positional information alone
should not help (the intruder is randomly distributed across the
sentence), suggests that other architectural biases are at
work. Intriguingly, a preliminary comparison of untrained BiLSTM-max
and human subjects on the SOMO sentences evaluated by both reveals
that, whereas humans have a bias towards finding sentences acceptable
(62\% sentences are rated as untampered with, vs.~48\% ground-truth
proportion), the model has a strong bias in the opposite direction (it
rates 83\% of the sentences as modified). A cursory look at
contrasting errors confirms, unsurprisingly, that those made by humans
are perfectly justified, while model errors are opaque. For example,
the sentence ``I didn't come here to \emph{reunite}
(orig.~\emph{undermine}) you'' seems perfectly acceptable in its
modified form, and indeed subjects judged it as such, whereas
untrained BiLSTM-max ``correctly'' rated it as a modified
item. Conversely, it is difficult to see any clear reason for the
latter tendency to rate perfectly acceptable originals as modified. We
leave a more thorough investigation to further work.  See similar
observations on the effectiveness of untrained ConvNets in vision by
\newcite{Ulyanov:etal:2017}.

\paragraph{Probing task comparison} A good encoder, such as
NMT-trained BiLSTM-max, shows generally good performance across
probing tasks. At one extreme,  performance is
not particularly high on the surface tasks, which might be an indirect
sign of the encoder extracting ``deeper'' linguistic properties. At
the other end, performance is still far from the human bounds on
TreeDepth, BShift, SOMO and CoordInv. The last 3 tasks ask if a
sentence is syntactically or semantically anomalous. This is a
daunting job for an encoder that has not been explicitly trained on
acceptability, and it is interesting that the best models are, at
least to a certain extent, able to produce reasonable anomaly
judgments. The asymmetry between the difficult TreeDepth and easier
TopConst is also interesting. Intuitively, TreeDepth requires more
nuanced syntactic information (down to the deepest leaf of the tree)
than TopConst, that only requires identifying broad chunks.

\insertalongtraining

Figure \ref{fig:along_training} reports how probing task accuracy
changes in function of encoder training epochs. The figure shows that
NMT probing performance is largely independent of target language,
with strikingly similar development patterns across French, German and
Finnish. Note in particular the similar probing accuracy curves in
French and Finnish, while the corresponding BLEU scores (in lavender)
are consistently higher in the former language. For both NMT and
SkipThought, WC performance keeps increasing with epochs. For the other tasks, we
observe instead an early flattening of the NMT probing curves, while
BLEU performance keeps increasing. Most strikingly, SentLen
performance is actually \emph{decreasing}, suggesting again that, as a
model captures deeper linguistic properties, it will tend to forget
about this superficial feature. Finally, for the challenging SOMO
task, the curves are mostly flat, suggesting that what
BiLSTM-max is able to capture about this task is already encoded in
its architecture, and further training doesn't help much.

\paragraph{Probing vs.~downstream tasks}
Figure \ref{fig:correl_matrix} reports correlation between performance
on our probing tasks and the downstream tasks available in the
SentEval\footnote{\url{https://github.com/facebookresearch/SentEval}} 
suite \cite{conneau2018senteval}, which consists of classification (MR, CR, SUBJ, MPQA, SST2, SST5, TREC), natural language inference (SICK-E), semantic relatedness (SICK-R, STSB), paraphrase detection (MRPC) and semantic textual similarity (STS 2012 to 2017) tasks.
Strikingly, WC is significantly positively correlated with all
downstream tasks. This suggests that, at least for current models, the
latter do not require extracting particularly abstract knowledge from
the data. Just relying on the \emph{words} contained in the input
sentences can get you a long way. Conversely, there is a significant
negative correlation between SentLen and most downstream tasks. The
number of words in a sentence is not informative about its linguistic
contents. The more models abstract away from such information, the
more likely it is they will use their capacity to capture more
interesting features, as the decrease of the SentLen curve along training (see Figure \ref{fig:along_training}) also suggests. CoordInv and, especially, SOMO, the tasks
requiring the most sophisticated semantic knowledge, are those that
positively correlate with the largest number of downstream tasks after
WC. We observe intriguing asymmetries: SOMO correlates with the SICK-E
sentence entailment test, but not with SICK-R, which is about modeling
sentence relatedness intuitions. Indeed, logical entailment requires
deeper semantic analysis than modeling similarity judgments. TopConst
and the number tasks negatively correlate with various similarity and
sentiment data sets (SST, STS, SICK-R). This might expose biases in
these tasks: SICK-R, for example, deliberately contains sentence pairs
with opposite voice, that will have different constituent structure
but equal meaning \cite{Marelli:etal:2014a}. It might also mirrors
genuine factors affecting similarity judgments (e.g., two sentences
differing only in object number are very similar). Remarkably, TREC
question type classification is the downstream task correlating with
most probing tasks. Question classification is certainly an outlier
among our downstream tasks, but we must leave a full understanding of
this behaviour to future work (this is exactly the sort of analysis
our probing tasks should stimulate).

\insertcorrelationmatrix

%% file: related_work.tex
\section{Related work}
\label{sec_related}

% Several recent papers share our interest in probing the knowledge
% contained in linguistic embeddings.
\newcite{Adi:etal:2017} introduced SentLen, WC and a word order test,
focusing on a bag-of-vectors baseline, an autoencoder and 
skip-thought (all trained on the same data used for the probing
tasks). We recast their tasks so that they only require a sentence
embedding as input (two of their tasks also require word embeddings,
polluting sentence-level evaluation), we extend the evaluation to more
tasks, encoders and training objectives, and we relate performance on
the probing tasks with that on downstream
tasks. \newcite{Shi:etal:2016} also use 3 probing tasks, including
Tense and TopConst. It is not clear that they controlled for the same
factors we considered (in particular, lexical overlap and sentence
length), and they use much smaller training sets, limiting
classifier-based evaluation to logistic regression.  Moreover, they
test a smaller set of models, focusing on machine translation.% We
% did not re-implement their \emph{voice} task since we did not see a
% way to generate cases where the passive voice is not given away by
% trivial surface cues.

\newcite{Belinkov:etal:2017}, \newcite{Belinkov:etal:2017b} and \newcite{Dalvi:etal:2017} are also
interested in understanding the type of linguistic knowledge encoded
in sentence and word embeddings, but their focus is  on
word-level morphosyntax and lexical semantics, and specifically on NMT encoders and
decoders. %Loic: similar to the probing tasks but using a contrastive approach instead of a classification approach 
\newcite{Sennrich:2017} also focuses on NMT systems, and proposes a contrastive test to assess how they handle various linguistic phenomena. % (e.g., subject-verb agreement). 
Other work explores the linguistic behaviour
of recurrent networks and related models by using visualization,
input/hidden representation deletion techniques or by looking at the
word-by-word behaviour of the network
\citep[e.g.,][]{Nagamine:etal:2015,Hupkes:etal:2017,Li:etal:2016,Linzen:etal:2016,Kadar:etal:2017,Li:etal:2017}. These
methods, complementary to ours, are not
agnostic to encoder architecture, and cannot be used
for general-purpose cross-model evaluation.

%Loic: this is less directly related, but it goes towards the "inspection" of the network/representations
% it not only answers the "what" but also the "how" and "where"
% \newcite{Nagamine:etal:2015} proposed a deep analysis of the node activations in different layers of the network 
% in order to understand how the phonemes are encoded in their ASR system. This work is complementary to ours since 
% it corresponds to an intrinsic analysis of the network which explains how and where the phonemes are encoded in the network.

Finally, \newcite{Conneau:etal:2017} propose a large-scale, multi-task
evaluation of sentence embeddings, focusing entirely on 
downstream tasks.%  Our work complements theirs, by focusing on
% probing tasks.

% Finally, to the best of our knowledge, we are the first to make our
% probing task battery publicly available, for others to benchmark their models on.

%% file: conclusion.tex
\section{Conclusion}

We introduced a set of tasks probing the linguistic knowledge of
sentence embedding methods. Their purpose is not to encourage the
development of \emph{ad-hoc} models that attain top performance on them,
but to help exploring what information is captured by different
pre-trained encoders.

We performed an extensive linguistic evaluation of
modern sentence encoders. Our results suggest
that the encoders are capturing a wide range of properties,
well above those captured by a set of strong baselines. We further
uncovered interesting patterns of correlation between the probing
tasks and more complex ``downstream'' tasks, and presented a set of
intriguing findings about the linguistic properties of various
embedding methods. For example, we found that Bag-of-Vectors is
surprisingly good at capturing sentence-level properties, thanks to
redundancies in natural linguistic input. We showed that different encoder architectures trained with the same objective with similar performance can result in different embeddings, pointing out the importance of the architecture prior for sentence embeddings. In particular, we found that BiLSTM-max
embeddings are already capturing interesting linguistic knowledge before
training, and that, after training, they detect semantic
acceptability without having been exposed to anomalous sentences before. We hope that our publicly available probing task set
will become a standard benchmarking tool of the linguistic properties
of new encoders, and that it will stir research towards a better
understanding of what they learn.

In future work, we would like to extend the probing tasks to other languages (which should be relatively easy, given that they are automatically generated), investigate how multi-task training affects probing task performance and leverage our probing tasks to find more linguistically-aware universal encoders.
% AC: Are we saying too much here?

%We would further like to combine the probing tasks with
%model introspection. For example, we would like to explore whether
%there are activations, in a certain decoder, that correlate, say, with
%syntactic tasks but not with semantic ones. Finally, we would like to
%evaluate both how multi-task training affects probing task
%performance, and whether the probing tasks themselves could be used
%as auxiliary tasks for more linguistically-aware training of universal
%encoders.

%% file: appendix.tex
\newpage
\section{Appendix}
\label{sec_appendix}

\subsection*{Amazon Mechanical Turk survey}

Subjects were recruited through the standard Amazon Mechanical Turk interface.\footnote{\url{https://www.mturk.com/}} 
We created independent surveys for the SOMO, CoordInv and BShift tasks. We asked  subjects to identify which sentences were acceptable and which were anomalous/inverted. Participants were restricted to those based in an English-speaking country.

To maximize annotation quality, we created a control set. Two authors annotated 200 random sentences from each task in a blind pretest. Those sentences on which they agreed were included in the control set.

We collected at least 10 judgments per sentence, for 1k random sentences from each task. We only retained judgments by subjects that rated at least 10 control sentences with accuracy of at least 90\%. After filtering, we were left with averages of 2.5, 2.9 and 12 judgments per sentence for SOMO, CoordInv and BShift, respectively. Responses were aggregated by majority voting, before computing the final accuracies.

We did not record any personal data from subjects, and we only used the judgments in aggregated format to produce the estimated human upper bounds reported in our tables.

\subsection*{Further training details}

\paragraph{Encoder training} For seq2seq tasks, after hyper-parameter
tuning, we chose 2-layer LSTM decoders with 512 hidden units. For NLI,
we settled on a multi-layer perceptron with 100 hidden units. As is
now common in NMT, we apply Byte Pair Encoding (BPE)
\citep{Sennrich:2017} to target sentences only, with 40k codes (see
Table 1 in the main text for examples of transformed target
sentences). We tune dropout rate and input embedding size, picking
1024 for BiLSTMs and 512 for Gated ConvNets. We use the Adam optimizer
for BiLSTMs and SGD with momentum for Gated ConvNets (after Adam gave
very poor results). The encoder representation is fed to the decoder
at every time step. For model selection on the validation sets, we use
BLEU score\footnote{MOSES multi-bleu.perl script
  \cite{koehn2007moses}} for NMT and AutoEncoder,
perplexity for SkipThought and accuracy for Seq2Tree and NLI.

Table \ref{table:test_results} reports test set performance of the various architectures on the original training tasks. For NMT and Seq2Tree, we left out two random sets of 10k sentences from the training data for dev and test. The NLI dev and test sets are the ones of SNLI. Observe how results are similar for the three encoders, while, as discussed in the main text, they differ in terms of the linguistic properties their sentence embeddings are capturing. The last row of the table reports BLEU scores for our BiLSTM architecture trained with attention, showing that the architecture is on par with current NMT models, when attention is introduced. For comparison, our attention-based model obtains 37 BLEU score on the standard WMT'14 En-Fr benchmark.

\inserttestresults

\paragraph{Probing task training} The probing task results reported in
the main text are obtained with a MLP that uses the Sigmoid nonlinearity,
which we found to perform better than Tanh. We tune the $L^2$
regularization parameter, the number of hidden states (in [50, 100,
200]) and the dropout rate (in [0, 0.1, 0.2]) on the validation set of
each probing task. Only for WC, which has significantly more
output classes (1000) than the other tasks, we report Logistic
Regression results, since they were consistently better.

\subsection*{Logistic regression results}

Logistic regression performance approximates MLP performance (compare
Table \ref{table:probing_logreg_results} here to Table 2 in the main
text). This suggests that most linguistic properties can be extracted
with a linear readout of the embeddings. Interestingly, if we focus on a good model-training combination, such as BiLSTM-max trained on French NMT, the tasks where the improvement from logistic
regression to MLP is relatively large ($>$3\%) are those arguably
requiring the most nuanced linguistic knowledge (TreeDepth, SOMO,
CoordInv).
\insertembsize

\insertlogregprobingtable

\subsection*{Downstream task results}
   
\insertdownstreamtable

We evaluate our architecture+training method combinations on the
downstream tasks from the SentEval
toolkit.\footnote{\url{https://github.com/facebookresearch/SentEval}} See
documentation there for the tasks, that range from subjectivity
analysis to question-type classification, to paraphrase detection and
entailment. Also refer to the SentEval page and to
\citet{Conneau:etal:2017} for the specifics of training and figures of
merit for each task. In all cases, we used as input our pre-trained
embeddings without fine-tuning them to the tasks. Results are reported
in Table \ref{table:downstream_results}.

We replicate the finding of Conneau and colleagues about the
effectiveness of the BiLSTM architecture with max pooling, that has
also a slight edge over GatedConvNet (an architecture they did not
test). As for the probing tasks, we again notice that BiLSTM-max is
already effective without training, and more so than the alternative
architectures.

Interestingly, we also confirm Conneau et al.'s finding that NLI is
the best source task for pre-training, despite the fact that, as we
saw in the main text (Table 2 there), NMT pre-training leads to models
that are capturing more linguistic properties. As they observed for downstream tasks, increasing the embedding dimension while adding capacity to the model is beneficial (see Figure \ref{fig:embsize}) also for probing tasks in the case of NLI. However, it does not seem to boost the performance of the NMT En-Fr encoder.

Finally, the table also shows results from the literature recently obtained
with various state-of-the-art general-purpose encoders, namely: SkipThought with layer
normalization \citep{ba2016layer}, InferSent (BiLSTM-max as trained on NLI by
Conneau et al.) and MultiTask \citep{subramanian2018learning}. A
comparison of these results with ours confirms that we are testing
models that do not lag much behind the state of the art.